\documentclass{article}

\PassOptionsToPackage{numbers, compress}{natbib}
 \usepackage[preprint]{neurips_2026}

\usepackage[utf8]{inputenc} 
\usepackage[T1]{fontenc}    
\usepackage{hyperref}       
\usepackage{url}            
\usepackage{booktabs}       
\usepackage{amsfonts}       
\usepackage{nicefrac}       
\usepackage{microtype}      
\usepackage{xcolor}         

\usepackage{amsmath}
\usepackage{graphicx} 
\usepackage{wrapfig}
\usepackage{booktabs}
\usepackage{makecell}
\usepackage{array}
\usepackage{multirow}
\usepackage{xcolor}
\usepackage{arydshln}
\newcolumntype{C}[1]{>{\centering\arraybackslash}m{#1}}

\title{StrLoRA: Towards Streaming Continual Visual Instruction Tuning for MLLMs}

\author{%
Chang Che\textsuperscript{1}\quad
Ziqi Wang\textsuperscript{1}\quad
Hui Ma\textsuperscript{1}\quad
Cheems Wang\textsuperscript{2}\quad
Zenglin Shi\textsuperscript{1}\thanks{Corresponding author: zenglin.shi@hfut.edu.cn} \\
\textsuperscript{1}Hefei University of Technology \\
\textsuperscript{2}Tsinghua University
}

\begin{document}

\maketitle

\begin{abstract}
Continual Visual Instruction Tuning (CVIT) enables Multimodal Large Language Models to incrementally acquire new abilities. However, existing CVIT methods operate under a restrictive task-incremental setting, where each training phase corresponds to a single, predefined task. This does not reflect real-world conditions, where data arrives as a continuous stream of interleaved and dynamically evolving tasks.
To bridge this gap, we introduce Streaming CVIT (StrCVIT), a more general and realistic setting where models learn from a stream of data chunks containing a dynamic mixture of tasks. In StrCVIT, a model must simultaneously acquire new abilities, reinforce recurring abilities, and mitigate forgetting. Existing CVIT methods fail here as they cannot reliably distinguish or adapt to the heterogeneous task samples within each chunk.
We therefore propose StrLoRA, a regularized two-stage expert routing framework. StrLoRA first performs task-aware expert selection using the textual instruction to activate a sparse subset of relevant experts, reducing cross-task interference. It then applies token-wise expert weighting within this subset, where contribution weights are computed via cross-modal attention between local visual tokens and the global instruction representation. To maintain stability across the non-stationary stream, a routing-stability regularization aligns current routing distributions with a historical exponential moving average reference.
Extensive experiments on a newly developed StrCVIT benchmark show that StrLoRA substantially outperforms existing methods, effectively enhancing model's abilities from continuously evolving data streams. The code is available at \url{https://github.com/chanceche/StrCVIT}.
\end{abstract}

\section{Introduction}
Multimodal Large Language Models (MLLMs)~\cite{bai2025qwen3vl, gemmateam2025gemma3technicalreport,wang2025internvl3} have demonstrated remarkable capabilities that extend beyond their text-only counterparts~\cite{grattafiori2024llama,yang2025qwen3}. Typically developed through a multi-stage pipeline~\cite{bai2025qwen3vl, wang2025internvl3}, these models undergo initial pretraining on large-scale image-text pairs followed by visual instruction tuning to align responses with human intent. For real-world deployment, MLLMs must continuously acquire new vision-language abilities to adapt to evolving requirements. Prior work formalizes this process as Continual Visual Instruction Tuning (CVIT)~\cite{chen2024coin,wang2025smolora,che_2026}, where data arrives in discrete stages, each assumed to correspond to a single, predefined task. However, this task-incremental setting is a key limitation: real-world data typically arrives as a continuous, interleaved stream of multiple evolving tasks, not as neatly segregated units.

\begin{figure*}[t]
    \centering
    \includegraphics[width=0.99\textwidth]{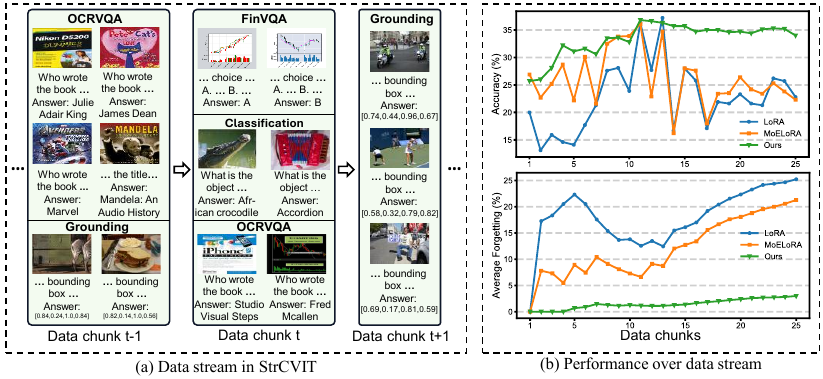}
    \caption{ (a) In StrCVIT, data arrives as a single-pass stream of chunks, each containing a dynamic mixture of interleaved tasks whose composition and proportions shift over time.  (b) Performance over the stream exhibits strong oscillations and suffers abrupt drops, while our proposed StrLoRA maintains stable accuracy and consistently low forgetting.}
    \label{fig1}
\end{figure*}

To bridge this gap, we introduce Streaming CVIT (StrCVIT), a more realistic paradigm where a model learns from a single-pass stream of data chunks. As shown in Fig.~\ref{fig1}~(a), each chunk contains a dynamic mixture of interleaved tasks; the number of active tasks, their composition, and sample proportions can all change over time. This setting requires a model to simultaneously acquire new abilities from emerging tasks, reinforce recurring abilities, and mitigate forgetting of capabilities absent from the current chunk. As shown in Fig.~\ref{fig1}~(b), this induces significant performance instability, with accuracy oscillating and suffering abrupt drops, a critical barrier to reliable, continual learning. We therefore measure forgetting continuously as the relative drop from peak historical performance, capturing the stability of acquired knowledge throughout the stream.

Existing CVIT methods are ill-suited for StrCVIT. They often employ Mixture-of-Experts (MoE) architectures~\cite{fedus2022switch,shazeer2017outrageously} with task-specific routing~\cite{che_2026,wang2025smolora,song2026kss}, which presumes each new stage introduces a single task. In StrCVIT, where chunks contain heterogeneous samples from multiple tasks, this strategy fails. The router cannot produce stable, discriminative routing weights, leading to homogeneous expert utilization, where diverse samples activate experts in highly similar ways (Fig.~\ref{fig2}~(a)), undermining parameter isolation and causing catastrophic interference.

We identify that textual instructions provide a stable, discriminative signal for task semantics, unlike visually entangled features (Fig.~\ref{fig2}~(b)). Leveraging this insight, we propose StrLoRA, a regularized two-stage expert routing framework. First, it performs task-aware expert selection, using the textual instruction to activate a sparse, relevant subset of LoRA experts, thereby reducing interference. Second, within this subset, a token-wise expert weighting module computes fine-grained contribution weights by attending to both local visual tokens and the global instruction. To combat instability in the data stream, we introduce a routing-stability regularization that aligns current routing decisions with a historical exponential moving average, ensuring smooth evolution of expert allocations. Our design thus decouples stable, text-guided expert filtering from adaptive, cross-modal token weighting.

We construct a comprehensive benchmark to simulate the StrCVIT scenario. Extensive experiments demonstrate that StrLoRA significantly outperforms existing methods, effectively enhancing multimodal abilities from evolving data streams while maintaining consistently low forgetting.

Our main contributions are summarized as follows:
\begin{itemize}
    \setlength{\itemsep}{0pt}
    \item We introduce StrCVIT, a new and realistic continual learning setting for MLLMs that models data as a single-pass stream of interleaved and dynamically evolving tasks.
    \item We diagnose the homogeneous expert utilization problem in StrCVIT and propose StrLoRA, a regularized two-stage expert routing framework that decouples task-aware expert selection from token-wise expert weighting, stabilized by a regularization loss.
    \item We construct a StrCVIT benchmark and experimentally validate that StrLoRA achieves superior performance, effectively enhancing abilities from a non-stationary stream while minimizing forgetting.
\end{itemize}

\section{Streaming Continual Visual Instruction Tuning}
\paragraph{StrCVIT.}
Given a pre-trained MLLM $f_{\theta}$, StrCVIT continually enhances its capabilities by fine-tuning on a single-pass data stream, which is
formulated as a sequence of data chunks $\mathcal{S} = (d_1, d_2, \dots, d_t, d_{t+1}, \dots)$. At each step $t$, the incoming data chunk is defined as $d_t=\{(x_i^{\text{vis}},x_i^{\text{text}},y_i^{\text{gt}})\}_{i=1}^{n_t}$ with $n_t$ samples, where each sample consists of a visual input, a textual instruction, and a ground-truth response.
The samples in $d_t$ are drawn from a mixture of multiple interleaved task distributions:
\begin{equation}
P_t = \sum_{m=1}^{M} \pi_t^{(m)} P^{(m)},
\label{distribution}
\end{equation}
where $M$ denotes the total number of task distributions observed in the stream so far, and $\pi_t^{(m)} \in [0,1]$ denotes the mixture weight of task distribution $P^{(m)}$ at time step $t$, with $\sum_{m=1}^{M}\pi_t^{(m)} = 1$. The number, composition, and sample proportions of interleaved tasks can all change dynamically across chunks. Under this formulation, the overall optimization objective over the streaming process up to time step $T$ is:
\begin{equation}
\min_{\theta_1, \dots, \theta_T}
\sum_{t=1}^{T}
\mathbb{E}_{(x^{\text{vis}},x^{\text{text}},y^{\text{gt}})\sim d_t}
\left[
\mathcal{L}\left(f_{\theta_t}(x^{\text{vis}},x^{\text{text}}),y^{\text{gt}}\right)
\right],
\quad \text{s.t.} \;\; \theta_t \leftarrow \theta_{t-1},
\label{eq_loss}
\end{equation}
where $\theta_t$ denotes the model parameters obtained after updating $\theta_{t-1}$ on $d_t$. 

\paragraph{Revisiting CVIT.}
The CVIT setting~\citep{che_2026,wang2025smolora} presents the model with a sequence of tasks $\mathcal{T}=\{\tau_1,\tau_2,\dots,\tau_n\}$ to acquire new vision-language abilities incrementally, where each task $\tau_k$ is associated with a dataset $\mathcal{D}_k=\{(x_{i}^{\text{vis}},x_{i}^{\text{text}},y^{\text{gt}}_i)\}_{i=1}^{n_k}$. This setting can be regarded as a restricted form of Eq.~\eqref{distribution} in which each task corresponds to a single previously unseen distribution. At step $k$, the mixture reduces to $\pi_k^{(k)}=1$ and $\pi_k^{(m)}=0$ for all $m \neq k$. The corresponding optimization objective is:
\begin{equation}
\min_{\theta_1, \dots, \theta_n}
\sum_{k=1}^{n}
\mathbb{E}_{(x^{\text{vis}},x^{\text{text}},y^{\text{gt}})\sim\mathcal{D}_k}
\left[
\mathcal{L}\left(f_{\theta_k}(x^{\text{vis}},x^{\text{text}};\tau_k),y^{\text{gt}}\right)
\right],
\quad \text{s.t.} \;\; \theta_k \leftarrow \theta_{k-1},
\end{equation}
where $f_{\theta_k}(\cdot;\tau_k)$ denotes the model that can explicitly or implicitly leverage the task identity $\tau_k$ for task-specific adaptation.

\paragraph{Key Challenges in StrCVIT.}
In CVIT, the model focuses on incrementally learning a single new task at each stage while mitigating forgetting of prior tasks.  StrCVIT, in contrast, requires learning from multiple, interleaved tasks at once. The model must therefore acquire new abilities from emerging tasks, reinforce recurring abilities from known tasks, and mitigate forgetting of abilities that are absent from the current chunk.
This fundamental difference necessitates distinct evaluation paradigms. While CVIT typically measures forgetting as the performance drop on previous tasks after learning new ones, StrCVIT requires continuous assessment: because performance may fluctuate with each chunk, we measure forgetting at every step relative to the best historical accuracy, capturing the stability of all acquired capabilities throughout the entire learning stream. Detailed evaluation metrics are provided in Section~\ref{sec:eval_metrics}.

\section{StrLoRA} 
\begin{figure*}[t]
    \centering
    \includegraphics[width=\textwidth]{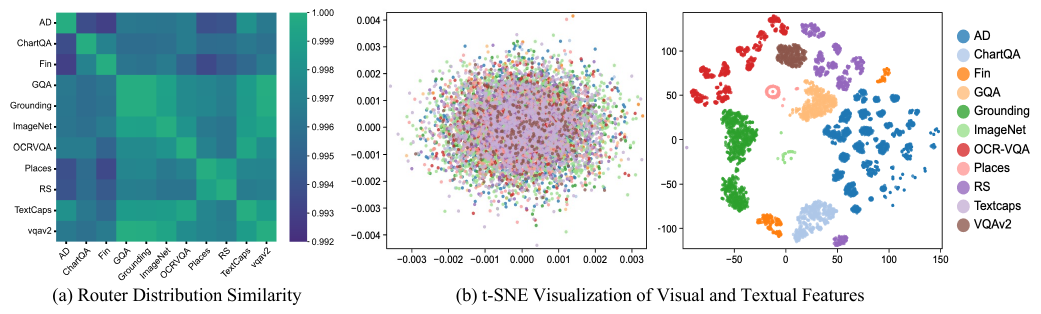}
    \caption{(a) CKA similarity~\citep{kornblith2019similarity} heatmap of StrCVIT, showing that standard MoE routing leads to homogeneous expert activation across samples from different tasks. (b) t-SNE visualization~\citep{van2008visualizing} of visual and textual features, showing that textual embeddings form well-separated clusters and provide a more reliable signal for task discrimination than overlapping visual features.}
    \label{fig2}
\end{figure*}

Existing CVIT methods typically employ a LoRA-based MoE architecture to reduce interference among different tasks. Specifically, the architecture maintains $N$ LoRA blocks as experts, each parameterized by a low-rank matrix product $BA$ inserted into the frozen pretrained weight $W_0$, and a task-specific router that assigns expert weights for each newly arriving task.
However, in StrCVIT, each data chunk consists of interleaved samples from multiple tasks that dynamically shift over time. This prevents the router from relying on task-specific routing, making it difficult to generate stable and discriminative routing patterns for heterogeneous samples. As illustrated in Fig.~\ref{fig2}(a), the router tends to produce highly similar expert-weight distributions across diverse samples, leading to homogeneous expert utilization that ultimately undermines effective expert specialization.

To address the unique challenges of StrCVIT, we propose StrLoRA, a regularized two-stage expert routing framework. First, it performs task-aware expert selection using the textual instruction to activate a sparse, relevant subset of experts, reducing cross-task interference. Then, it conducts token-wise expert weighting within that subset, assigning contribution weights by attending to both local visual tokens and the global instruction. To maintain stability in streaming data, a routing-stability regularization aligns current allocations with a historical EMA reference. This design leverages textual stability for coarse filtering and cross-modal synergy for fine-grained adaptation, making it robust for interleaved, evolving task streams.

\subsection{Task-Aware Expert Selection}
In multimodal instruction data, the textual instruction explicitly defines the task to be performed on the accompanying image, encapsulating its underlying semantics. We investigate whether this textual information can provide a robust and discriminative signal for expert routing. A dimensionality reduction analysis, shown in Fig.~\ref{fig2}(b), reveals a clear disparity: text embeddings form well-separated clusters corresponding to different tasks, while visual features exhibit significant overlap. This critical observation demonstrates that the instruction text offers a stable and reliable signal for distinguishing task types, unlike the more ambiguous visual modality.

Driven by this insight, we use the text embedding to perform task-aware expert selection, which selects the subset of experts to activate for a given sample. This approach directly addresses the homogeneous routing problem by leveraging the most discriminative available signal.
Let $\mathbf{x}^{\mathrm{text}}_i = \frac{1}{L_i^{\mathrm{text}}}\sum_{m=1}^{L_i^{\mathrm{text}}} \mathbf{t}_i^m\in \mathbb{R}^{d_\mathrm{e}}$ denote the global text embedding for sample $i$ obtained by mean pooling over $L_i^{\mathrm{text}}$ textual token embeddings, where $\mathbf{t}_i^m \in \mathbb{R}^{d_\mathrm{e}}$ is the $m$-th textual token embedding, and $d_\mathrm{e}$ is the embedding dimension. A linear router computes a probability distribution over all $N$~experts:
\begin{equation}
p_{ij}=
\frac{\exp\!\big((W_g \mathbf{x}^{\mathrm{text}}_i)_j\big)}
{\sum_{j'=1}^{N}\exp\!\big((W_g \mathbf{x}^{\mathrm{text}}_i)_{j'}\big)},
\quad
\mathbf p_i = [p_{ij}]_{j=1}^{N}, 
\end{equation}
where $W_g \in \mathbb{R}^{N \times d_\mathrm{e}}$ is the router projection matrix. $p_{ij}$ is the routing probability assigned to expert $j$ for sample $i$, and $\mathbf{p}_i$ is the full distribution over all experts. 
Then, we define the expert subset $\mathcal{S}_i = \operatorname{Top-K}(\mathbf{p}_i)$ by retaining the $\operatorname{K}$ experts with the highest routing probabilities for sample $i$. Since the router operates on the global text embedding, samples with similar intents activate consistent expert subsets across the stream, reducing interference from unrelated task intents.

\subsection{Token-Wise Expert Weighting}
Guided by the analysis that the textual instruction provides a clear, task-discriminative signal, we first use the text embedding to perform task-aware expert selection. This determines a relevant subset of experts $\mathcal S_i$ to activate for a given input sample $i$, effectively filtering out irrelevant experts and mitigating interference.

However, applying uniform expert weights to all tokens within a sample is suboptimal, as different visual regions correspond to distinct features whose relevance depends on the instructional context. Therefore, we introduce a second stage of token-wise expert weighting to compute adaptive contribution weights within the selected expert subset $\mathcal S_i$.

Formally, for the $l$-th token of sample $i$ with hidden state $h_i^l \in \mathbb{R}^{d_{\mathrm{in}}}$, we project it and the global text embedding $\mathbf{x}_i^{\mathrm{text}}$ into a shared $D$-dimensional routing space using learnable matrices $W_Q$ and $W_K$, respectively. Each expert $E_j$ is associated with a learnable feature vector $\mathbf{e}_j \in \mathbb{R}^{D}$. We denote the collection of expert feature vectors by $W_E=\{\mathbf{e}_j\}_{j=1}^{N}$.
 The routing logit for expert $E_j$ (where $j \in \mathcal S_i$) is calculated as:
\begin{equation}
z_{ij}^l =
\begin{cases}
\dfrac{(W_Qh_i^l) \cdot \big((W_K\mathbf{x}^{\mathrm{text}}_i) \odot \mathbf{e}_j\big)}{\sqrt{D}}, & j \in \mathcal S_i, \\[6pt]
-\infty, & j \notin \mathcal S_i.
\end{cases}
\label{z_logits}
\end{equation}
where $\odot$ denotes the Hadamard product.
The key term $W_K \mathbf{x}_i^{\mathrm{text}} \odot \mathbf{e}_j$ produces an instruction-adapted expert representation, dynamically modulating the expert's profile based on the current task. The inner product then scores the relevance of this adapted expert to the specific visual token $h_i^l$. The logits are normalized over the selected subset $\mathcal S_i$ to obtain the final contribution weights:
\begin{equation}
s_{ij}^l =
\frac{\exp(z_{ij}^l)}
{\sum_{j'\in \mathcal S_i}\exp(z_{ij'}^l)}.
\label{s_logits}
\end{equation}
For experts not in $\mathcal S_i$, the weight $s_{ij}^l$ is zero. The final adapted representation $\hat{h}_i^l$ combines the frozen pre-trained weight $W_0$ with a weighted sum of the activated LoRA experts $(B_j, A_j)$:
\begin{equation}
\hat{h}_i^l = W_0h_i^l + \sum_{j\in\mathcal S_i} s_{ij}^{l} \cdot B_j A_jh_i^{l},
\end{equation}
where $W_0 \in \mathbb{R}^{d_{\mathrm{out}} \times d_{\mathrm{in}}}$, $B_j \in \mathbb{R}^{d_{\mathrm{out}} \times r}$ and $A_j \in \mathbb{R}^{r \times d_{\mathrm{in}}}$ with $r \ll \min(d_{\mathrm{in}}, d_{\mathrm{out}})$.

\begin{figure*}[t]
    \centering
    \includegraphics[width=\textwidth]{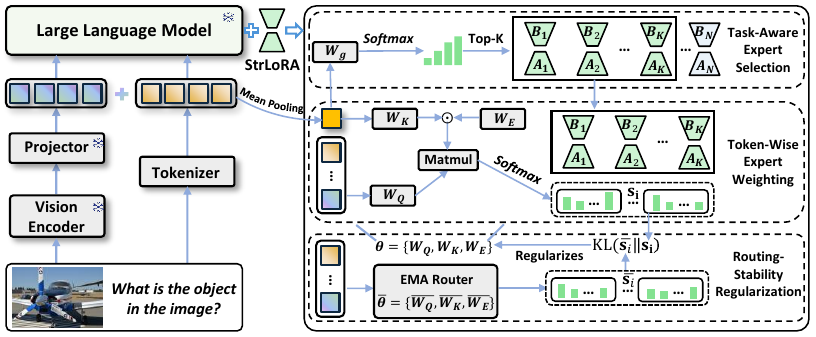}
    \caption{StrLoRA framework. StrLoRA first performs task-aware expert selection to activate a sparse subset of relevant experts. It then conducts token-wise expert weighting within the selected subset, assigning contribution weights. To maintain stability in streaming data, a routing-stability regularization aligns current allocations with a historical EMA reference.}
    \label{fig3}
\end{figure*}

\subsection{Routing-Stability Regularization} 
As the data stream evolves, rapid shifts in task distribution can cause the token-level routing module to overfit to the current chunk, deviating from previously learned allocation patterns. This instability undermines consistent expert specialization and can lead to performance fluctuations. To promote stable yet adaptive expert weight allocation, we introduce a routing-stability regularization loss. This loss encourages the token-level routing distribution to evolve smoothly within the currently activated expert subset, preserving historical allocation knowledge while remaining adaptable to new data.
We maintain an exponential moving average~(EMA) $\bar{\theta} = \{\bar{W_Q}, \bar{W_K}, \bar{W_E}\}$ of the routing parameters, updated after each training step $u$ as:
$\bar{\theta}^{(u)} = \beta\, \bar{\theta}^{(u-1)} + (1-\beta)\, \theta^{(u)},$
where $\beta \in [0,1)$ is the momentum coefficient. The EMA parameters serve as a stabilized reference of past routing~behavior.

For a given sample $i$ with selected expert subset $\mathcal S_i$, we compute the historical routing weights $\bar{s}^{l}_{ij}$ using $\bar{\theta}$ via the same routing equations (Eq.~\eqref{z_logits} and Eq.~\eqref{s_logits}). The regularization term is then defined as the KL divergence~\cite{kullback1951information} from the historical distribution to the current one, summed over all tokens:
\begin{equation}
\mathcal{L}_{\mathrm{reg}} =
\sum_{l=1}^{L_i}
\mathrm{KL}\!\left(
\bar{\mathbf{s}}_{i}^{\,l}
\,\middle\|\,
\mathbf{s}_{i}^{\,l}
\right),
\end{equation}
where $L_i$ is the number of tokens in sample $i$, and $\mathrm{KL}(\bar{\mathbf{s}}_{i}^{\,l} \| \mathbf{s}_{i}^{\,l}) = \sum_{j\in\mathcal S_i} \bar{s}_{ij}^{l} \log \frac{\bar{s}_{ij}^{l}}{s_{ij}^{l}}
$.

By penalizing large deviations from the smoothed historical routing, $\mathcal{L}_{\mathrm{reg}}$ prevents abrupt re-allocation of expert contributions, thus preserving routing consistency across the learning stream. The final training objective combines this with the standard instruction-tuning loss $\mathcal{L}$ defined in Eq.~\eqref{eq_loss}:
\begin{equation}
\mathcal{L}_{\mathrm{total}} = \mathcal{L} + \lambda\mathcal{L}_{\mathrm{reg}},
\end{equation}
where $\lambda$ balances task performance and routing stability. This design allows the model to adapt continually without forgetting how experts were productively assigned in the past, a key capability for learning robustly from non-stationary, interleaved task streams.

\section{Experimental Results}
\label{section:experiment}
\subsection{Benchmark Setup}
\label{section:benchmark}
\paragraph{Streaming Data Construction.}
To approximate a realistic data stream, we integrate vision-language datasets from a range of domains, including ImageNet~\citep{deng2009imagenet}, Places365~\citep{zhou2017places}, AD~\citep{sima2024drivelm}, Fin~\citep{wang2025harmonious}, OCRVQA~\citep{mishra2019ocr}, Grounding~\citep{chen2024coin}, TextCaps~\citep{sidorov2020textcaps}, VQAv2~\citep{goyal2017making}, RS~\citep{lobry2020rsvqa}, ChartQA~\citep{masry2022chartqa}, and GQA~\citep{hudson2019gqa}. At each time step $t$, we construct a data chunk by randomly sampling 2{,}000 samples from a randomly selected subset of 1 to 11 datasets, where the proportion contributed by each dataset varies and all samples are randomly shuffled. In this way, each chunk may contain samples from multiple interleaved distributions, and the stream composition changes over time. Detailed data stream statistics are presented in Appendix~\ref{append:data_stream}.

\paragraph{Evaluation Metrics.}
\label{sec:eval_metrics}
Inspired by~\citep{chaudhry2019tiny}, we measure forgetting to quantify the performance degradation of the model on each dataset over the stream. Specifically, let $a_t^m$ denote the accuracy score on the $m$-th dataset after the model processes the $t$-th data chunk. The forgetting on dataset $m$ at time step $t$ is defined as:
\begin{equation}
\mathrm{F}_t^m = \max\left(0, \frac{\max_{1 \le j < t} a_j^m - a_t^m}{\max_{1 \le j < t} a_j^m}\right).
\label{formula_F}
\end{equation}

This measures the relative performance drop on dataset $m$ with respect to its historical best previous state. To evaluate the model's performance on each individual dataset across the entire stream, we compute Average Performance (AP) \citep{che_2026} and Average Forgetting (AF):
\begin{equation}
\mathrm{AP}_t^m = \frac{1}{t}\sum_{j=1}^{t} a_j^m,
\qquad
\mathrm{AF}_t^m = \frac{1}{t}\sum_{j=1}^{t} \mathrm{F}_j^m.
\label{formula_per_dataset}
\end{equation}

To comprehensively evaluate the model's overall performance across all datasets, we introduce Mean Average Performance (MAP) and Mean Average Forgetting (MAF):
\begin{equation}
\mathrm{MAP}_t = \frac{1}{M}\sum_{m=1}^{M} \mathrm{AP}_t^m,
\qquad
\mathrm{MAF}_t = \frac{1}{M}\sum_{m=1}^{M} \mathrm{AF}_t^m.
\label{formula_overall}
\end{equation}

\subsection{Experimental Setup}
\label{section:experiment_setup}
\paragraph{Compared Methods.}
\label{compared_methods}
We compare our method with existing approaches applicable to StrCVIT. \textbf{LoRA}~\citep{hu2022lora} fine-tunes the model using a single shared LoRA module across all tasks. We also include classical CL approaches such as \textbf{EWC}~\citep{kirkpatrick2017overcoming} which constrains updates on parameters important to previous tasks, and \textbf{Replay}~\citep{chaudhry2019tiny} stores or generates past samples for rehearsal during training on new data. In addition, we evaluate methods designed for CVIT, including \textbf{MoELoRA}~\citep{chen2024coin}, which employs a mixture-of-experts module over LoRA experts, and \textbf{SMoLoRA}~\citep{wang2025smolora},  which introduces separate routing mechanisms for visual understanding and instruction following across different tasks. Finally, \textbf{Zero-shot} evaluates the base model without any fine-tuning and serves as a lower-bound reference.

\paragraph{Base Models.}
To comprehensively evaluate performance across different model architectures and scales, we conduct experiments on the StrCVIT benchmark using InternVL3.5-8B-Pretrained~\citep{wang2025internvl3} and Gemma3-4B-PT~\citep{gemmateam2025gemma3technicalreport} as base models, with additional results on InternVL3.5-4B-Pretrained presented in Appendix~\ref{append:additional_eval}.

\begin{table*}[t]
\centering
\caption{The evaluated results (\%) over 25 data chunks for StrCVIT across different base models. Higher AP/MAP is better, while lower AF/MAF is better.}

\label{tab:main_results}
\setlength{\tabcolsep}{3.4pt}
\renewcommand{\arraystretch}{1.2}
\resizebox{\textwidth}{!}{%
\begin{tabular}{
>{\centering\arraybackslash}m{1.9cm}
!{\vrule width 0.6pt}
*{8}{cc}
cc
!{\vrule width 0.6pt}
cc
}
\toprule
\multirow{2}{*}{\textbf{Method}}
& \multicolumn{2}{c}{Places365}
& \multicolumn{2}{c}{RS}
& \multicolumn{2}{c}{AD}
& \multicolumn{2}{c}{VQAV2}
& \multicolumn{2}{c}{Fin}
& \multicolumn{2}{c}{Grounding}
& \multicolumn{2}{c}{OCRVQA}
& \multicolumn{2}{c}{TextCaps}
& \multicolumn{2}{c}{ImageNet}
& \multicolumn{2}{c}{Overall} \\
\cmidrule(r){2-3}
\cmidrule(lr){4-5}
\cmidrule(lr){6-7}
\cmidrule(lr){8-9}
\cmidrule(lr){10-11}
\cmidrule(lr){12-13}
\cmidrule(lr){14-15}
\cmidrule(lr){16-17}
\cmidrule(lr){18-19}
\cmidrule(l){20-21}
& AP$\uparrow$ & AF$\downarrow$
& AP$\uparrow$ & AF$\downarrow$
& AP$\uparrow$ & AF$\downarrow$
& AP$\uparrow$ & AF$\downarrow$
& AP$\uparrow$ & AF$\downarrow$
& AP$\uparrow$ & AF$\downarrow$
& AP$\uparrow$ & AF$\downarrow$
& AP$\uparrow$ & AF$\downarrow$
& AP$\uparrow$ & AF$\downarrow$
& MAP$\uparrow$ & MAF$\downarrow$ \\
\specialrule{0.08em}{0.12em}{0.08em}
\multicolumn{21}{l}{\textbf{InternVL3.5-8B-Pretrained}} \\
\midrule
Zero-shot
& 4.70 & -- & 53.40 & -- & 23.00 & -- & 28.70 & -- & 63.70 & -- & 24.90 & -- & 25.70 & -- & 49.68 & -- & 26.10 & -- & 33.32 & -- \\
LoRA
& 22.78 & 25.22 & 68.73 & 1.26 & 32.20 & 4.11 & 69.20 & 1.28 & 74.23 & 0.79 & 57.40 & 3.99 & \textbf{54.36} & 2.13 & 83.89 & 0.96 & 73.48 & 3.19 & 59.59 & 4.77 \\
EWC
& 24.45 & 21.62 & 68.36 & 1.00 & 31.19 & \textbf{2.02} & 70.15 & 0.83 & 72.97 & 0.79 & \textbf{59.40} & \textbf{2.39} & 52.34 & 3.71 & 83.05 & 1.17 & 66.25 & 3.48 & 58.68 & 4.11 \\
Replay
& 21.62 & 28.85 & 67.96 & 2.09 & 31.42 & 4.41 & \textbf{70.25} & 0.81 & 74.31 & 1.22 & 57.34 & 3.32 & 54.25 & 2.03 & 84.25 & 0.91 & 72.75 & 3.49 & 59.35 & 5.24 \\
MoELoRA
& 26.14 & 21.27 & 69.10 & 1.62 & 33.94 & 4.51 & 70.14 & \textbf{0.72} & 74.23 & 0.84 & 55.30 & 6.90 & 53.67 & \textbf{1.86} & 84.41 & \textbf{0.65} & 74.78 & 3.81 & 60.19 & 4.69 \\
SMoLoRA
& 17.34 & 20.21 & 64.29 & 2.97 & 29.27 & 4.31 & 69.27 & 1.55 & 71.52 & 1.04 & 43.93 & 8.23 & 40.71 & 4.00 & 75.83 & 5.54 & 49.98 & 3.80 & 51.35 & 5.74 \\
StrLoRA 
& \textbf{33.37} & \textbf{2.99} & \textbf{69.65} & \textbf{0.75} & \textbf{36.27} & 3.92 & 70.11 & 1.13 & \textbf{75.95} & \textbf{0.38} & 57.36 & 2.53 & 53.00 & 3.29 & \textbf{84.48} & 0.74 & \textbf{76.49} & \textbf{1.46} & \textbf{61.85} & \textbf{1.91} \\
\specialrule{0.08em}{0.12em}{0.08em}
\multicolumn{21}{l}{\textbf{Gemma3-4B-PT}} \\
\midrule
Zero-shot
& 13.50 & -- & 45.50 & -- & 18.30 & -- & 14.70 & -- & 1.50 & -- & 0.00 & -- & 6.00 & -- & 0.09 & -- & 32.20 & -- & 14.64 & -- \\
LoRA
& 32.08 & 6.92 & 58.88 & 4.68 & 24.49 & 20.06 & 55.67 & 2.21 & 67.05 & 3.57 & 5.16 & 17.90 & 56.22 & 4.10 & 70.79 & 3.67 & 63.57 & \textbf{3.14} & 48.21 & 7.36 \\
EWC
& 31.67 & 7.17 & 58.89 & 4.80 & 24.57 & 21.72 & 55.53 & 2.50 & 66.85 & 3.85 & 5.00 & 18.37 & 56.42 & 3.74 & 70.84 & 3.14 & 62.82 & 4.31 & 48.07 & 7.73 \\
Replay
& 31.36 & 7.12 & 59.96 & 5.41 & 24.31 & 14.76 & \textbf{56.04} & 1.78 & 68.90 & 1.82 & 4.49 & 30.88 & 57.92 & \textbf{2.38} & 70.77 & 3.17 & 63.93 & 3.16 & 48.63 & 7.83 \\
MoELoRA
& \textbf{32.70} & \textbf{5.36} & 58.30 & 5.17 & 27.68 & 11.45 & 55.15 & 3.27 & 66.72 & 4.71 & 4.98 & 26.21 & 56.04 & 2.95 & \textbf{71.26} & 2.78 & 63.58 & 3.63 & 48.49 & 7.28 \\
SMoLoRA
& 29.05 & 9.60 & 58.48 & 4.62 & 21.51 & 17.68 & 54.49 & 2.76 & 67.58 & 2.81 & 3.42 & 27.16 & 50.45 & 2.71 & 69.46 & \textbf{2.51} & 62.74 & 3.87 & 46.36 & 8.19 \\
StrLoRA 
& 32.36 & 6.66 & \textbf{60.52} & \textbf{3.30} & \textbf{32.63} & \textbf{8.02} & 55.00 & \textbf{0.91} & \textbf{72.20} & \textbf{0.87} & \textbf{6.20} & \textbf{12.88} & \textbf{58.32} & 3.64 & 68.45 & 8.71 & \textbf{73.54} & 4.15 & \textbf{51.03} & \textbf{5.46} \\
\bottomrule
\end{tabular}%
}
\vspace{-0.5cm}
\end{table*}

\subsection{Main Results}
\paragraph{StrCVIT Setting.} We evaluate StrLoRA against all baselines over 25 streaming data chunks across different model architectures and scales. As shown in Table~\ref{tab:main_results}, on InternVL3.5-8B-Pretrained, all compared methods exhibit substantial forgetting on at least one dataset. 
Although SMoLoRA mitigates inter-task interference in CVIT through task-specific routing, it struggles with mixed data streams spanning multiple tasks. In contrast, StrLoRA achieves strong accuracy on most datasets, notably improving AP on Places365 from 26.14 with MoELoRA to 33.37 while reducing AF from 21.27 to 2.99, and reaches superior overall performance with an MAP of 61.85 and an MAF of 1.91. These results demonstrate that StrLoRA continuously strengthens multimodal capabilities on incoming data while maintaining consistently low forgetting in StrCVIT.
On the cross-architecture setting with Gemma3-4B-PT, StrLoRA also achieves stable performance and outperforms all baselines, highlighting the robustness of StrLoRA across different architectures.

\paragraph{CVIT Setting.} We evaluate StrLoRA on the CoIN benchmark~\citep{chen2024coin} to demonstrate that our method can also generalize to the standard CVIT setting. Following the CoIN setting, we use LLaVA-v1.5-7B~\citep{liu2024visual} as the backbone. We compare StrLoRA with multiple public CVIT baselines under the instruction original setting. As shown in Table~\ref{tab:coin_eval}, StrLoRA outperforms all compared methods on the overall performance metrics.
\begin{table*}[ht]
\vspace{-0.5cm}
\centering
\caption{Results (\%) on the CoIN benchmark. MFT denotes mean fine-tuning accuracy across tasks, MFN denotes mean final accuracy across the eight tasks, MAA denotes mean average accuracy over all tasks, and BWT measures backward transfer, where negative values indicate forgetting.}
\label{tab:coin_eval}
\setlength{\tabcolsep}{3.2pt}
\renewcommand{\arraystretch}{1.0}
\resizebox{\textwidth}{!}{%
\begin{tabular}{c cccccccccccc}
\toprule
\textbf{Method} &
ScienceQA$\uparrow$ &
TextVQA$\uparrow$ &
ImageNet$\uparrow$ &
GQA$\uparrow$ &
VizWiz$\uparrow$ &
Grounding$\uparrow$ &
VQAv2$\uparrow$ &
OCR-VQA$\uparrow$ &
MFT$\uparrow$ &
MFN$\uparrow$ &
MAA$\uparrow$ &
BWT$\uparrow$ \\
\midrule
MoELoRA~\citep{chen2024coin} & 72.91 & 48.82 & 9.58 & 42.55 & 36.70 & 3.40 & 48.62 & 63.87 & 62.63 & 40.81 & 45.00 & -21.83 \\
BranchLoRA~\citep{zhang2025enhancing} & 68.24 & 40.18 & 24.60 & 41.40 & 49.83 & 15.94 & 51.23 & 62.14 & 65.18 & 44.20 & 49.94 & -20.98 \\
KSS-MoE~\citep{song2026kss} & 68.73 & 54.79 & 57.33 & 52.11 & 42.86 & 15.37 & 60.69 & 47.68 & 61.20 & 49.95 & 59.99 & -11.25 \\
SMoLoRA~\citep{wang2025smolora} & 79.75 & 47.99 & \textbf{93.88} & 42.44 & 46.03 & 17.80 & 53.64 & 57.69 & 63.82 & 54.90 & 63.01 & -8.92 \\
ProgLoRA~\citep{yu-etal-2025-progressive} & 74.84 & 51.83 & 83.90 & 49.93 & \textbf{53.87} & 31.19 & 62.71 & 64.44 & 65.68 & 59.09 & 62.38 & -6.59 \\
StrLoRA & \textbf{83.59} & \textbf{61.87} & 59.49 & \textbf{61.04} & 38.74 & \textbf{59.80} & \textbf{64.80} & \textbf{65.50} & \textbf{68.42} & \textbf{61.85} & \textbf{67.90} & \textbf{-6.57} \\
\bottomrule
\end{tabular}%
}
\end{table*}

\begin{figure*}[t]
    \centering
    \includegraphics[width=\textwidth]{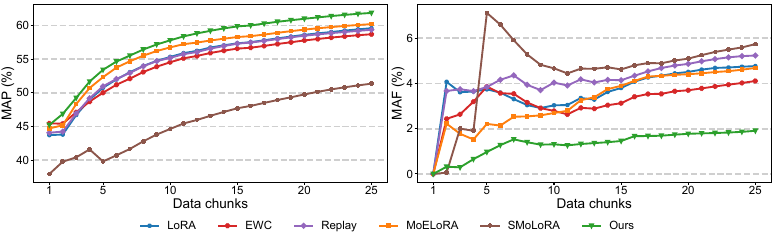}
    \vspace{-0.6cm}
    \caption{Overall performance over the data stream on InternVL3.5-8B-Pretrained.}
    \label{performance}
    \vspace{-0.2cm}
\end{figure*}

\paragraph{Visualization of Overall Performance.}
We further visualize the evolution of MAP and MAF on InternVL3.5-8B-Pretrained as the data stream progresses. As shown in Fig.~\ref{performance}, StrLoRA maintains a clear lead on the accuracy curve while keeping forgetting consistently the lowest across the entire data stream. 
This trend highlights that StrLoRA not only reaches stronger final performance, but also preserves a more stable adaptation trajectory during streaming learning.

\begin{figure*}[h]
    \centering
    \includegraphics[width=\textwidth]{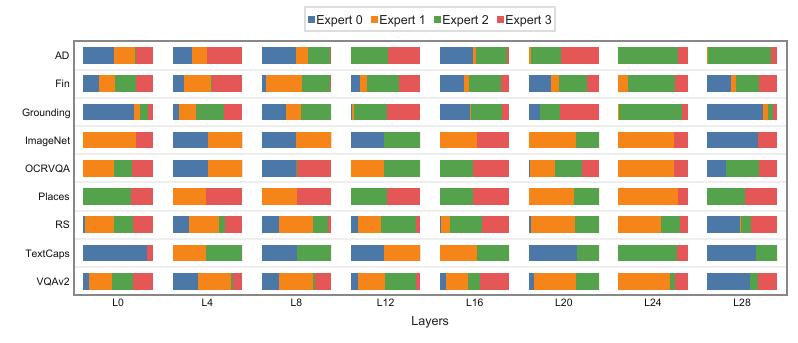}
    \vspace{-0.8cm}
    \caption{Expert activation proportions across tasks. Each stacked bar represents the activation proportion of each expert, averaged over all samples.}
    \vspace{-0.5cm}
    \label{fig:expert_activation}
\end{figure*}

\paragraph{Expert Activation Analysis.}
To further explore whether StrLoRA can stably distinguish heterogeneous samples and achieve expert specialization, we visualize the expert activation proportions across layers for each task in Fig.~\ref{fig:expert_activation}. 
 The results show that samples from different tasks activate different subsets of experts with varying proportions, exhibiting distinct expert preferences across layers. This suggests that StrLoRA effectively partitions the parameter space to reduce interference among heterogeneous samples drawn from multiple interleaved task distributions.

\subsection{Ablation Study}
We conduct ablation studies to examine the contribution of each component in StrLoRA using InternVL3.5-8B-Pretrained. 
Specifically, $p_{ij}$ denotes sample-level expert weights. $s_{ij}^l$ denotes token-level expert weights. $\mathcal{L}_{\mathrm{reg}}$ denotes the routing-stability regularization term. We also include MoELoRA setting, where none of these components are applied. 
As shown in Table~\ref{tab:ablation_modular}, using $p_{ij}$ alone improves the MAP from 60.19 to 60.77 and reduces MAF from 4.69 to 3.24, but the AF on Places365 remains high at 16.05, suggesting that uniform weighting is insufficient to prevent forgetting.
Using $s_{ij}^l$ alone also improves MAP and MAF, but the improvement is limited without $p_{ij}$ for Top-K expert selection to separate heterogeneous samples.
Combining $p_{ij}$ and $s_{ij}^l$ further improves MAP to 61.58, demonstrating that sample-level expert selection and token-level weight assignment are complementary. 
With all components enabled, StrLoRA achieves the best overall performance, with 61.85 MAP and MAF of 1.91. Notably, on Places365, AF drops from 21.27 to 2.99, confirming that StrLoRA maintains strong accuracy with consistently low forgetting across the training process. Additional ablation studies are presented in Appendix~\ref{app:hyperparameter_ablation}.

\begin{table*}[t]
\centering
\caption{Ablation study of the components within
StrLoRA on InternVL3.5-8B.}
\label{tab:ablation_modular}
\setlength{\tabcolsep}{3.4pt}
\renewcommand{\arraystretch}{1.2}
\resizebox{\textwidth}{!}{%
\begin{tabular}{
c c c
!{\vrule width 0.6pt}
*{8}{cc}
cc
!{\vrule width 0.6pt}
cc
}
\toprule
\multicolumn{3}{c}{\textbf{Component}}
& \multicolumn{2}{c}{Places365}
& \multicolumn{2}{c}{RS}
& \multicolumn{2}{c}{AD}
& \multicolumn{2}{c}{VQAV2}
& \multicolumn{2}{c}{Fin}
& \multicolumn{2}{c}{Grounding}
& \multicolumn{2}{c}{OCRVQA}
& \multicolumn{2}{c}{TextCaps}
& \multicolumn{2}{c}{ImageNet}
& \multicolumn{2}{c}{Overall} \\
\cmidrule(r){1-3}
\cmidrule(r){4-5}
\cmidrule(lr){6-7}
\cmidrule(lr){8-9}
\cmidrule(lr){10-11}
\cmidrule(lr){12-13}
\cmidrule(lr){14-15}
\cmidrule(lr){16-17}
\cmidrule(lr){18-19}
\cmidrule(lr){20-21}
\cmidrule(l){22-23}
$p_{ij}$ & $s_{ij}^l$ & $\mathcal{L}_{\mathrm{reg}}$
& AP$\uparrow$ & AF$\downarrow$
& AP$\uparrow$ & AF$\downarrow$
& AP$\uparrow$ & AF$\downarrow$
& AP$\uparrow$ & AF$\downarrow$
& AP$\uparrow$ & AF$\downarrow$
& AP$\uparrow$ & AF$\downarrow$
& AP$\uparrow$ & AF$\downarrow$
& AP$\uparrow$ & AF$\downarrow$
& AP$\uparrow$ & AF$\downarrow$
& MAP$\uparrow$ & MAF$\downarrow$ \\
\specialrule{0.08em}{0.12em}{0.08em}

$\times$ & $\times$ & $\times$
& 26.14 & 21.27 & 69.10 & 1.62 & 33.94 & 4.51 & 70.14 & 0.72 & 74.23 & 0.84 & 55.30 & 6.90 & 53.67 & 1.86 & 84.41 & 0.65 & 74.78 & 3.81 & 60.19 & 4.69 \\

$\checkmark$ & $\times$ & $\times$
& 28.91 & 16.05 & 70.18 & 2.22 & 34.72 & 3.37 & 70.48 & 0.60 & 76.54 & 0.82 & 57.88 & 3.85 & 49.89 & 2.12 & 83.71 & 1.32 & 74.59 & 1.49 & 60.77 & 3.54 \\

$\times$ & $\checkmark$ & $\times$
& 27.04 & 19.40 & 68.34 & 0.67 & 33.58 & 2.91 & 70.30 & 1.14 & 73.60 & 0.71 & 56.36 & 4.10 & 57.60 & 4.55 & 83.89 & 0.77 & 73.68 & 2.22 & 60.49 & 4.05 \\

$\times$ & $\checkmark$ & $\checkmark$
& 28.78 & 10.18 & 68.80 & 1.34 & 32.30 & 3.25 & 70.36 & 0.85 & 73.24 & 1.97 & 56.21 & 4.76 & 58.57 & 3.78 & 84.04 & 0.95 & 73.50 & 2.10 & 60.64 & 3.24 \\

$\checkmark$ & $\checkmark$ & $\times$
& 30.06 & 11.78 & 68.88 & 2.31 & 35.86 & 2.26 & 69.44 & 1.30 & 76.75 & 0.39 & 60.17 & 3.76 & 54.39 & 1.32 & 85.00 & 0.81 & 73.65 & 2.78 & 61.58 & 2.97 \\

$\checkmark$ & $\checkmark$ & $\checkmark$
& 33.37 & 2.99 & 69.65 & 0.75 & 36.27 & 3.92 & 70.11 & 1.13 & 75.95 & 0.38 & 57.36 & 2.53 & 53.00 & 3.29 & 84.48 & 0.74 & 76.49 & 1.46 & 61.85 & 1.91 \\

\bottomrule
\end{tabular}%
}
\vspace{-0.5cm}
\end{table*}

\section{Related Work}
\subsection{Online Continual Learning}
Beyond the traditional class-incremental continual learning setting, task-free continual learning \citep{aljundi2019task} introduces a task-free formulation in which task boundaries are unavailable during training. Building on this setting, Si-Blurry \citep{moon2023online} extends it to online class-incremental learning under a stochastic blurry scenario, where class distributions change over time. DCM \citep{ye2024online} further studies online task-free continual learning in a setting where class identities are unavailable and supervised signals may also be inaccessible during training. MiD-Blurry \citep{wang2026symmetric} introduces an online continual learning benchmark that combines multiple class distribution patterns with blurred temporal boundaries. 
Nevertheless, these studies are confined to classification-oriented continual learning with homogeneous vision-language tasks, and thus do not align the practical need for MLLMs to continually acquire and reinforce diverse multimodal capabilities. This mismatch in task scope and learning objectives makes their methods not directly applicable to StrCVIT. Although OASIS~\citep{lee2025oasis} introduces online sample selection for CVIT, it still operates under a task-incremental setting with a predefined task sequence, rather than a fully streaming scenario with continuously mixed multimodal tasks.

\subsection{CVIT Methods}
To mitigate inter-task interference in CVIT, a wide range of methods have been proposed under the task-incremental setting. BranchLoRA \citep{zhang2025enhancing} shares LoRA A matrices, learns task-specific B branches, and uses task-specific routers for each new task. ProgLoRA \citep{yu-etal-2025-progressive} maintains a progressive LoRA pool and trains a new LoRA block for each new task. CL-MoE \citep{huai2025cl} distinguishes task-shared and task-specific experts selected by dual routers and updates them with different momentum weights as new tasks arrive. SMoLoRA \citep{wang2025smolora}  introduces separate routing mechanisms for visual understanding and instruction following across different tasks. D-MoLE \citep{ge2025dynamic} dynamically allocates new task-specific LoRA experts across layers for each new task under a parameter budget. LiLoRA \citep{che_2026} dynamically expands lightweight task-specific LoRA modules for each new task. KSS-MoE \citep{song2026kss} adds a new task-specific expert for each new task and regularizes it to be distinct from previous experts. However, these methods rely on the setting that each incoming training data is a single predefined task to enable explicit task-specific module allocation, and thus cannot be directly applied to StrCVIT, where each incoming data chunk is drawn from multiple interleaved tasks.

\section{Conclusion}
In this paper, we introduce StrCVIT, a new and realistic continual learning setting for MLLMs that models data as a single-pass stream of interleaved and dynamically evolving tasks. 
This setting exposes a key limitation of existing CVIT methods, where standard MoE routing often leads to homogeneous expert utilization, weakened expert specialization, and increased cross-task interference.
To address this challenge, we propose StrLoRA, a regularized two-stage routing framework that combines task-aware expert selection, cross-modal token-wise expert weighting, and EMA-based routing-stability regularization. Extensive experiments on our proposed StrCVIT benchmark demonstrate that StrLoRA substantially outperforms existing methods, effectively enhancing the model’s abilities from continuously evolving data streams.


{\small
\bibliographystyle{IEEEtran}
\bibliography{references}
}

\clearpage
\appendix
\section{Limitations}
\label{app_limitations}
Due to the difficulty of collecting real-world MLLM data streams with naturally evolving and interleaved task distributions, we cannot directly obtain such streams at the required scale. Therefore, we use multiple existing vision-language datasets to construct sequential data chunks with varying task compositions and sample proportions to simulate the StrCVIT scenario.
Although this benchmark approximates practical distribution shifts, it may not fully capture all temporal dependencies, noise patterns, and user-driven changes that occur in real-world deployment.

\section{Broader Impact}
\label{app_impact}
StrCVIT focuses on continual acquisition and retention of multimodal capabilities in MLLMs trained on non-stationary data streams with interleaved task distributions. Its primary positive impact is to provide researchers with a structured setting to evaluate adaptation, forgetting, and stability under more realistic streaming conditions. The benchmark can also reveal transient performance drops that remain hidden in standard task-incremental CVIT evaluation, enabling more reliable assessment prior to downstream deployment.

A primary risk is that continual adaptation methods could be used to push deployed MLLMs toward biased, harmful, or overly narrow behavior as new data are collected. A further risk is that results obtained on simulated streams may be overinterpreted as guarantees for real-world deployment, where data quality, temporal dependencies, and user interactions are typically more complex. We mitigate these risks by framing StrCVIT as an evaluation setting, constructing streams from publicly available vision-language datasets, and emphasizing stability and forgetting analysis rather than uncontrolled online model updating.

\section{Detailed Data Stream Composition}
\label{append:data_stream}
\subsection{Dataset Statistics}
Table~\ref{tab:data_stream_composition} summarizes the datasets used in StrCVIT. The benchmark covers eleven tasks, ranging from general capabilities such as grounding, captioning, classification, and VQA to domain-specific scenarios including autonomous driving, finance, remote sensing, OCR, and chart understanding.
\begin{table}[ht]
\centering
\small
\caption{ Statistics of the collected datasets in the StrCVIT benchmark.}
\label{tab:data_stream_composition}
\renewcommand{\arraystretch}{1.45}
\setlength{\tabcolsep}{4pt}
\begin{tabular}{C{0.36\textwidth} C{0.2\textwidth} C{0.2\textwidth} C{0.14\textwidth}}
\Xhline{1.2pt}
\textbf{Task} & \textbf{Train Dataset} & \textbf{Test Dataset} & \makecell[c]{\textbf{Test Number}} \\
\Xhline{1.2pt}
\textbf{Grounding} &
{\renewcommand{\arraystretch}{1.1}
\begin{tabular}[c]{@{}c@{}}
RefCOCO\\[-1pt]
RefCOCO+\\[-1pt]
RefCOCOg
\end{tabular}} &
{\renewcommand{\arraystretch}{1.1}
\begin{tabular}[c]{@{}c@{}}
RefCOCO\\[-1pt]
RefCOCO+\\[-1pt]
RefCOCOg
\end{tabular}} &
1000 \\
\hdashline
\textbf{Image Captioning} & TextCaps & TextCaps & 3166 \\
\hdashline
\textbf{Image Classification} & ImageNet & ImageNet & 1000 \\
\hdashline
\textbf{Scene Classification} & Places365 & Places365 & 1000 \\
\hdashline
\textbf{Visual Question Answering~(VQA)} & VQAv2 & VQAv2 & 1000 \\
\hdashline
\textbf{Visual Reasoning VQA} & GQA & GQA & 1000 \\
\hdashline
\textbf{Autonomous Driving VQA~(AD)} & DriveLM & DriveLM & 1000 \\
\hdashline
\textbf{Financial VQA~(Fin)} & StockQA & StockQA & 1000 \\
\hdashline
\textbf{Remote Sensing VQA~(RS)} & RSVQA & RSVQA & 1000 \\
\hdashline
\textbf{OCR VQA} & OCRVQA & OCRVQA & 1000 \\
\hdashline
\textbf{Chart VQA} & ChartQA & ChartQA & 1250 \\
\Xhline{1.2pt}
\end{tabular}
\end{table}

\begin{figure*}[ht]
    \centering
    \includegraphics[width=0.95\textwidth]{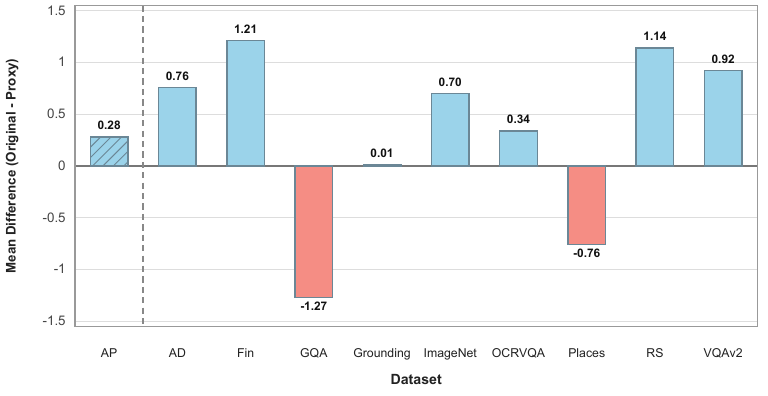}
    \caption{
    Mean accuracy differences between original and proxy test sets on data 001 to data 025. 
    }
    \label{fig:proxy_original}

\end{figure*}
\begin{figure*}[ht]
    \centering
    \includegraphics[width=0.9\textwidth]{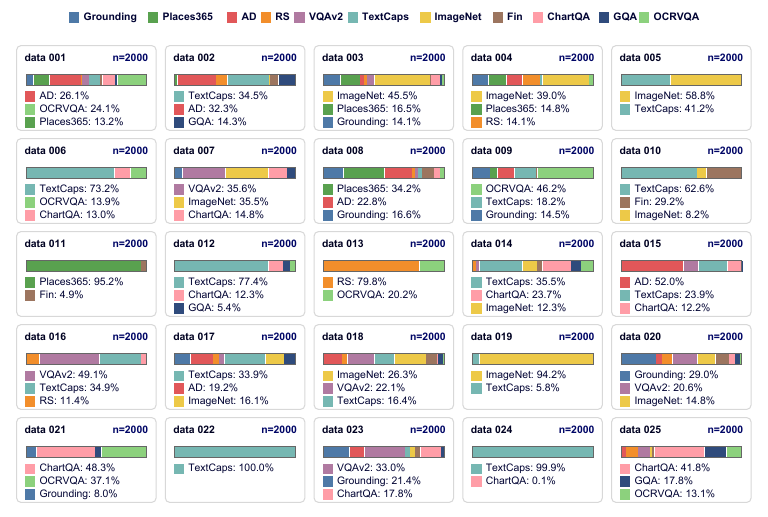}
    \caption{
    Composition of data chunks from data 001 to data 025. Each chunk contains 2,000 samples, and the stacked bars show the proportions of different datasets within each chunk.
    }
    \label{fig:dataset_composition}

\end{figure*}

\subsection{Proxy Test Sets}
For ImageNet, Places365, AD, Fin, OCRVQA, Grounding, VQAv2, RS, and GQA, we construct proxy test sets by stratified sampling from the original test sets according to their annotation distributions, reducing evaluation cost while preserving the evaluation trends of the full test sets. To validate the reliability of these proxy sets, we evaluate LoRA on both the original and proxy test sets using InternVL3.5-8B-Pretrained. As shown in Fig.~\ref{fig:proxy_original}, the proxy results closely track the original test results across data 001 to data 025, indicating that the proxy sets provide a faithful approximation for monitoring performance over the stream.

\subsection{Data Chunk Composition}
Fig.~\ref{fig:dataset_composition} shows the composition of each data chunk in the data stream. Notably, Places365 appears only in the early stage and no longer appears after data 011, reflecting a realistic non-stationary task distribution where some tasks may disappear from the incoming stream.

\section{Implementation Details}
\label{Implementation_details}
\subsection{Base Models}
\paragraph{InternVL3.5-8B-Pretrained.} InternVL3.5-8B-Pretrained is a multimodal InternVL model pretrained on large-scale vision-language pairs. The experiment checkpoint is available at~\url{https://huggingface.co/OpenGVLab/InternVL3_5-8B-Pretrained}.
\paragraph{InternVL3.5-4B-Pretrained.} InternVL3.5-4B-Pretrained is a multimodal InternVL model pretrained on large-scale vision-language pairs. The experiment checkpoint is available at~\url{https://huggingface.co/OpenGVLab/InternVL3_5-4B-Pretrained}.
\paragraph{Gemma3-4B-PT.} Gemma3-4B-PT is a pretrained multimodal Gemma 3 model pretrained on large-scale vision-language pairs. The experiment checkpoint is available at~\url{https://huggingface.co/google/gemma-3-4b-pt}.
\paragraph{LLaVA-v1.5-7B.} LLaVA-v1.5-7B is a multimodal model trained by visual instruction tuning on
top of a Vicuna language backbone and a CLIP vision encoder. The experiment checkpoint is available at~\url{https://huggingface.co/liuhaotian/llava-v1.5-7b}

\subsection{Training Setup}
\label{app:Training_Setup}
All experiments are conducted on a server with 2 Intel Xeon Platinum 8368 CPUs at 2.40GHz and 8 NVIDIA A100-SXM4-80GB GPUs. We implement training with PyTorch 2.9.1+cu128 and CUDA 12.8 under the ms-swift~\cite{zhao2025swift} 3.11.0.dev0 training framework. 
StrLoRA adapters are inserted into both the FeedForward Network layers and the self-attention layers of the LLM, while the projection layer between the vision encoder and the LLM is fully fine-tuned. We use the straight-through estimator to allow gradient flow through the discrete top-k selection.
The hyperparameters for StrLoRA are configured as follows: the number of LoRA experts is set to 4, the $\operatorname{Top-K}$ selection is 2, and each LoRA expert has a rank of 16. In the routing-stability regularization loss $\mathcal{L}_{\mathrm{reg}}$, the EMA momentum coefficient $\beta$ and the loss weight $\lambda$ are set to 0.99 and 0.1, respectively. We employ the Adam optimizer with a learning rate of $2\times10^{-5}$ and a batch size of 32. All data chunks are trained for a single epoch.

\section{Additional Experiments}
\label{append:additional_eval}

\subsection{Evaluation Under Different Stream Lengths}
We further present the evaluated results on the StrCVIT benchmark under 15 and 20 data chunks settings. As shown in Table~\ref{tab:additional_evaluation}, StrLoRA achieves consistently superior overall performance while maintaining low forgetting across all models.

\subsection{Evaluation on Datasets with Potential Pretraining Overlap}
ChartQA may overlap with the pretraining data of both InternVL3.5 and Gemma3-4B, and GQA may also overlap with the pretraining data of InternVL3.5, as suggested by the consistently high zero-shot scores. To isolate this factor, we present additional evaluations on these two datasets separately. As shown in Table~\ref{tab:additional_eval}, StrLoRA still maintains stable performance across the data stream. Since these datasets already exhibit strong zero-shot performance, the results mainly reflect whether continual tuning can preserve pretrained capabilities while adapting to the stream. StrLoRA remains competitive in AP and maintains low forgetting.

\begin{table*}[ht]
\centering
\caption{The evaluated results (\%) over 15 and 20 data chunks across different base models.}
\label{tab:additional_evaluation}
\setlength{\tabcolsep}{3.4pt}
\renewcommand{\arraystretch}{1.1}
\resizebox{\textwidth}{!}{%
\begin{tabular}{
>{\centering\arraybackslash}m{1.9cm}
!{\vrule width 0.6pt}
*{8}{cc}
cc
!{\vrule width 0.6pt}
cc
}
\toprule
\multirow{2}{*}{\textbf{Method}}
& \multicolumn{2}{c}{Places365}
& \multicolumn{2}{c}{RS}
& \multicolumn{2}{c}{AD}
& \multicolumn{2}{c}{VQAV2}
& \multicolumn{2}{c}{Fin}
& \multicolumn{2}{c}{Grounding}
& \multicolumn{2}{c}{OCRVQA}
& \multicolumn{2}{c}{TextCaps}
& \multicolumn{2}{c}{ImageNet}
& \multicolumn{2}{c}{Overall} \\
\cmidrule(r){2-3}
\cmidrule(lr){4-5}
\cmidrule(lr){6-7}
\cmidrule(lr){8-9}
\cmidrule(lr){10-11}
\cmidrule(lr){12-13}
\cmidrule(lr){14-15}
\cmidrule(lr){16-17}
\cmidrule(lr){18-19}
\cmidrule(l){20-21}
& AP$\uparrow$ & AF$\downarrow$
& AP$\uparrow$ & AF$\downarrow$
& AP$\uparrow$ & AF$\downarrow$
& AP$\uparrow$ & AF$\downarrow$
& AP$\uparrow$ & AF$\downarrow$
& AP$\uparrow$ & AF$\downarrow$
& AP$\uparrow$ & AF$\downarrow$
& AP$\uparrow$ & AF$\downarrow$
& AP$\uparrow$ & AF$\downarrow$
& MAP$\uparrow$ & MAF$\downarrow$ \\
\specialrule{0.08em}{0.12em}{0.08em}
\multicolumn{21}{l}{\textbf{InternVL3.5-8B-Pretrained 15 Chunks}} \\
\midrule
LoRA
& 22.81 & 16.10 & 65.66 & 1.66 & 31.99 & 4.11 & 67.90 & 1.53 & 72.49 & 0.54 & 53.85 & 4.57 & 53.03 & 1.02 & 82.40 & 0.94 & 65.91 & 3.74 & 57.34 & 3.80 \\
EWC
& 25.37 & 13.56 & 65.67 & 1.08 & 30.82 & \textbf{1.82} & 69.55 & 0.68 & 71.41 & 0.65 & \textbf{56.09} & 2.41 & 51.07 & 3.75 & 81.53 & 0.87 & 57.43 & 3.34 & 56.55 & 3.13 \\
Replay
& 22.05 & 20.49 & 64.38 & 2.80 & 31.26 & 4.10 & \textbf{69.97} & \textbf{0.66} & 72.84 & 0.69 & 54.09 & 3.60 & 52.96 & \textbf{0.85} & 82.64 & \textbf{0.54} & 65.35 & 3.57 & 57.28 & 4.15 \\
MoELoRA
& 27.70 & 12.72 & 65.83 & 2.20 & 33.90 & 3.81 & 69.53 & 0.72 & 72.54 & 0.75 & 52.17 & 7.51 & 51.94 & 2.42 & 82.88 & 0.63 & 67.95 & 4.29 & 58.27 & 3.90 \\
SMoLoRA
& 17.44 & 6.25 & 60.25 & 3.08 & 28.08 & 3.36 & 68.53 & 1.54 & 68.92 & 1.05 & 36.38 & 11.43 & 37.36 & 2.82 & 71.11 & 7.91 & 41.14 & 4.16 & 47.69 & 4.62 \\
Ours
& \textbf{32.42} & \textbf{1.37} & \textbf{67.45} & \textbf{0.93} & \textbf{34.87} & 4.20 & 69.35 & 1.40 & \textbf{73.52} & \textbf{0.27} & 54.78 & \textbf{1.82} & \textbf{53.07} & 1.46 & \textbf{82.97} & 0.63 & \textbf{70.23} & \textbf{1.04} & \textbf{59.85} & \textbf{1.46} \\
\specialrule{0.08em}{0.12em}{0.08em}
\multicolumn{21}{l}{\textbf{InternVL3.5-8B-Pretrained 20 Chunks}} \\
\midrule
LoRA
& 22.60 & 22.33 & 67.60 & 1.35 & 32.12 & 4.10 & 68.72 & 1.36 & 73.54 & 0.55 & 55.81 & 4.56 & \textbf{53.93} & \textbf{1.57} & 83.24 & 0.96 & 69.85 & 3.83 & 58.60 & 4.51 \\
EWC
& 24.95 & 18.16 & 67.34 & 1.03 & 31.13 & \textbf{1.63} & 70.01 & \textbf{0.66} & 72.26 & 0.80 & \textbf{57.88} & \textbf{2.56} & 51.97 & 3.54 & 82.41 & 1.13 & 61.94 & 3.81 & 57.76 & 3.70 \\
Replay
& 21.50 & 26.50 & 66.68 & 2.27 & 31.54 & 3.74 & \textbf{70.18} & 0.70 & 73.79 & 0.99 & 55.92 & 3.55 & 53.75 & 1.57 & 83.57 & 0.74 & 69.31 & 3.73 & 58.47 & 4.86 \\
MoELoRA
& 26.72 & 18.07 & 67.90 & 1.77 & 34.12 & 3.71 & 69.93 & 0.67 & 73.52 & 0.91 & 53.88 & 7.51 & 53.06 & 1.87 & 83.78 & 0.64 & 71.46 & 4.54 & 59.37 & 4.41 \\
SMoLoRA
& 17.73 & 13.76 & 62.71 & 3.10 & 28.99 & 3.37 & 69.06 & 1.45 & 70.17 & 1.27 & 40.25 & 9.41 & 39.88 & 2.60 & 74.12 & 6.31 & 44.84 & 4.62 & 49.75 & 5.10 \\
Ours
& \textbf{33.02} & \textbf{2.38} & \textbf{68.77} & \textbf{0.76} & \textbf{35.78} & 3.87 & 69.79 & 1.25 & \textbf{74.96} & \textbf{0.29} & 56.07 & 2.61 & 53.11 & 2.46 & \textbf{83.90} & \textbf{0.60} & \textbf{73.48} & \textbf{1.76} & \textbf{60.98} & \textbf{1.78} \\
\specialrule{0.08em}{0.12em}{0.08em}
\multicolumn{21}{l}{\textbf{InternVL3.5-4B-Pretrained 15 Chunks}} \\
\midrule
LoRA
& 29.32 & 5.48 & 63.64 & 4.77 & 34.47 & 1.68 & 68.70 & 1.75 & 74.07 & 0.98 & 58.20 & 2.09 & 56.09 & 3.09 & 81.38 & 0.51 & 63.28 & 2.93 & 58.79 & 2.59 \\
EWC
& 30.21 & 3.00 & 66.21 & \textbf{2.32} & 33.64 & 3.27 & 68.67 & 1.17 & 73.87 & 0.98 & 59.44 & 2.82 & \textbf{57.80} & 2.07 & 80.78 & 0.55 & 64.47 & \textbf{2.45} & 59.46 & 2.07 \\
Replay
& 29.96 & 4.34 & 65.05 & 4.69 & \textbf{34.62} & 1.88 & 68.41 & 1.69 & 74.47 & 0.87 & 56.42 & 5.81 & 57.55 & \textbf{1.09} & 81.81 & 0.44 & 63.57 & 3.38 & 59.09 & 2.69 \\
MoELoRA
& 30.04 & 4.92 & 65.56 & 2.89 & 34.45 & 2.19 & 68.73 & 0.89 & 73.18 & 0.80 & 58.97 & 2.23 & 57.26 & 2.82 & 80.91 & 0.87 & \textbf{66.29} & 2.55 & 59.49 & 2.24 \\
SMoLoRA
& 25.33 & 5.81 & 60.33 & 7.17 & 30.14 & 3.70 & \textbf{68.83} & 0.80 & 73.06 & 0.68 & 51.99 & 2.10 & 55.56 & 2.44 & 79.43 & \textbf{0.12} & 65.17 & 3.62 & 56.65 & 2.94 \\
Ours
& \textbf{31.21} & \textbf{2.99} & \textbf{66.29} & 2.56 & 33.61 & \textbf{1.36} & 68.58 & \textbf{0.59} & \textbf{75.72} & \textbf{0.62} & \textbf{60.05} & \textbf{1.23} & 57.09 & 1.42 & \textbf{82.00} & 0.33 & 65.39 & 5.33 & \textbf{59.99} & \textbf{1.82} \\
\specialrule{0.08em}{0.12em}{0.08em}
\multicolumn{21}{l}{\textbf{InternVL3.5-4B-Pretrained 20 Chunks}} \\
\midrule
LoRA
& 29.40 & 6.86 & 66.37 & 3.74 & 34.58 & 1.93 & 68.82 & 1.68 & 75.29 & 1.01 & 59.82 & 2.44 & 56.60 & 2.94 & 82.41 & 0.54 & 67.24 & 3.85 & 60.06 & 2.78 \\
EWC
& 30.28 & 5.34 & \textbf{68.41} & \textbf{1.93} & 33.76 & 3.56 & 68.90 & 1.14 & 75.18 & 0.98 & 60.62 & 3.37 & 57.88 & 2.58 & 81.70 & 0.81 & 68.78 & 2.75 & 60.61 & 2.50 \\
Replay
& 29.97 & 6.13 & 67.40 & 3.67 & \textbf{34.93} & 1.70 & 68.62 & 1.64 & 75.60 & 1.01 & 58.47 & 5.22 & \textbf{57.90} & 1.73 & 82.70 & 0.47 & 67.40 & 4.24 & 60.33 & 2.87 \\
MoELoRA
& 29.70 & 8.34 & 67.71 & 2.37 & 34.58 & 2.18 & 68.84 & 0.90 & 74.57 & 0.77 & 60.37 & 2.79 & 57.81 & 3.26 & 81.77 & 1.12 & \textbf{70.72} & \textbf{2.53} & 60.67 & 2.70 \\
SMoLoRA
& 25.89 & 9.05 & 63.75 & 5.57 & 31.45 & 3.26 & \textbf{68.98} & 1.09 & 74.29 & 0.72 & 54.16 & 2.26 & 55.97 & 3.00 & 80.71 & \textbf{0.33} & 69.22 & 3.92 & 58.27 & 3.24 \\
Ours
& \textbf{31.64} & \textbf{4.24} & 67.91 & 2.30 & 34.07 & \textbf{1.47} & 68.80 & \textbf{0.61} & \textbf{76.95} & \textbf{0.66} & \textbf{60.80} & \textbf{1.82} & 57.58 & \textbf{1.45} & \textbf{82.93} & 0.38 & 70.38 & 4.30 & \textbf{61.22} & \textbf{1.91} \\
\specialrule{0.08em}{0.12em}{0.08em}
\multicolumn{21}{l}{\textbf{Gemma3-4B-PT 15 Chunks}} \\
\midrule
LoRA
& \textbf{32.49} & \textbf{2.95} & 55.55 & 6.44 & 19.62 & 24.46 & 53.53 & 2.58 & 62.11 & 3.91 & 4.23 & \textbf{4.29} & 52.92 & 6.00 & 66.48 & 4.31 & 54.17 & \textbf{3.10} & 44.57 & 6.45 \\
EWC
& 32.01 & 3.78 & 55.75 & 6.34 & 19.27 & 25.14 & 53.57 & 2.91 & 61.78 & 4.30 & 4.21 & 5.08 & 53.17 & 5.56 & 66.24 & 4.39 & 52.48 & 6.29 & 44.28 & 7.09 \\
Replay
& 31.73 & 4.23 & 55.44 & 7.94 & 19.56 & 18.14 & \textbf{54.03} & 2.26 & 64.67 & 2.12 & 3.11 & 34.68 & 55.45 & 3.58 & 67.23 & \textbf{3.05} & 54.42 & 3.14 & 45.07 & 8.79 \\
MoELoRA
& 32.37 & 3.55 & 54.81 & 7.05 & 23.27 & 8.20 & 52.97 & 4.53 & 61.70 & 6.40 & \textbf{4.47} & 13.21 & 52.96 & 3.86 & 67.03 & 3.42 & 53.25 & 4.77 & 44.76 & 6.11 \\
SMoLoRA
& 29.26 & 3.59 & 55.05 & 6.49 & 15.75 & 15.01 & 52.07 & 3.36 & 62.77 & 3.36 & 3.06 & 14.15 & 44.81 & \textbf{3.36} & 64.28 & 3.27 & 53.45 & 3.60 & 42.28 & 6.24 \\
Ours
& 31.97 & 5.44 & \textbf{58.11} & \textbf{4.87} & \textbf{33.10} & \textbf{6.00} & 53.99 & \textbf{0.96} & \textbf{70.73} & \textbf{0.71} & 3.73 & 15.69 & \textbf{56.46} & 4.34 & \textbf{68.69} & 7.17 & \textbf{67.37} & 5.03 & \textbf{49.35} & \textbf{5.58} \\
\specialrule{0.08em}{0.12em}{0.08em}
\multicolumn{21}{l}{\textbf{Gemma3-4B-PT 20 Chunks}} \\
\midrule
LoRA
& 32.61 & 4.39 & 57.52 & 5.40 & 22.57 & 20.73 & 54.95 & 2.14 & 65.31 & 3.56 & 4.06 & 16.80 & 54.81 & 4.96 & 68.91 & 3.93 & 58.61 & 3.61 & 46.59 & 7.28 \\
EWC
& 32.33 & 4.40 & 57.68 & 5.34 & 22.94 & 21.03 & 54.86 & 2.51 & 65.00 & 3.95 & 4.05 & 17.40 & 54.96 & 4.49 & 68.93 & 3.57 & 57.73 & 5.20 & 46.50 & 7.54 \\
Replay
& 32.01 & 4.59 & 58.24 & 6.19 & 22.59 & 14.23 & \textbf{55.21} & 1.99 & 67.34 & 1.87 & 4.30 & 27.95 & 56.82 & 2.93 & 69.12 & 3.30 & 58.89 & \textbf{3.52} & 47.17 & 7.40 \\
MoELoRA
& \textbf{33.10} & \textbf{3.25} & 56.86 & 5.99 & 26.85 & 7.56 & 54.35 & 3.49 & 64.86 & 5.29 & 4.04 & 26.01 & 54.78 & 3.39 & \textbf{69.42} & 3.05 & 58.65 & 3.90 & 46.99 & 6.88 \\
SMoLoRA
& 29.50 & 6.29 & 57.19 & 5.33 & 19.99 & 13.58 & 53.55 & 2.87 & 65.62 & 3.20 & 2.88 & 26.17 & 48.10 & \textbf{2.75} & 67.44 & \textbf{2.75} & 57.56 & 4.33 & 44.65 & 7.47 \\
Ours
& 32.27 & 6.07 & \textbf{59.60} & \textbf{3.89} & \textbf{32.93} & \textbf{6.92} & 54.65 & \textbf{0.89} & \textbf{71.62} & \textbf{0.84} & \textbf{5.23} & \textbf{12.96} & \textbf{57.53} & 4.01 & 68.47 & 8.22 & \textbf{70.71} & 4.98 & \textbf{50.33} & \textbf{5.42} \\
\bottomrule
\end{tabular}%
}

\end{table*}

\begin{table*}[ht]
\centering
\caption{Additional evaluation on GQA and ChartQA across different base models.}
\label{tab:additional_eval}
\scriptsize
\setlength{\tabcolsep}{2.7pt}
\renewcommand{\arraystretch}{1.0}
\begin{tabular}{c *{12}{c}}
\toprule[0.7pt]
\multirow{3}{*}{\textbf{Method}}
& \multicolumn{4}{c}{\textbf{Gemma3-4B-PT}}
& \multicolumn{4}{c}{\textbf{InternVL3.5-4B-Pretrained}}
& \multicolumn{4}{c}{\textbf{InternVL3.5-8B-Pretrained}} \\
\cmidrule(lr){2-5}
\cmidrule(lr){6-9}
\cmidrule(lr){10-13}
& \multicolumn{2}{c}{GQA}
& \multicolumn{2}{c}{ChartQA}
& \multicolumn{2}{c}{GQA}
& \multicolumn{2}{c}{ChartQA}
& \multicolumn{2}{c}{GQA}
& \multicolumn{2}{c}{ChartQA} \\
\cmidrule(lr){2-3}
\cmidrule(lr){4-5}
\cmidrule(lr){6-7}
\cmidrule(lr){8-9}
\cmidrule(lr){10-11}
\cmidrule(lr){12-13}
& AP$\uparrow$ & AF$\downarrow$
& AP$\uparrow$ & AF$\downarrow$
& AP$\uparrow$ & AF$\downarrow$
& AP$\uparrow$ & AF$\downarrow$
& AP$\uparrow$ & AF$\downarrow$
& AP$\uparrow$ & AF$\downarrow$ \\
\midrule[0.45pt]
Zero-shot
& 4.00 & -- & 32.80 & --
& 57.50 & -- & 54.32 & --
& 60.40 & -- & \textbf{57.76} & -- \\
LoRA
& 43.84 & 4.23 & 33.27 & 1.53
& 60.84 & \textbf{0.92} & 53.98 & 0.55
& 61.88 & 2.09 & 57.43 & 0.76 \\
EWC
& 43.84 & 4.23 & 33.29 & 1.08
& 60.83 & 1.16 & 53.98 & 0.55
& 61.96 & \textbf{0.80} & 57.33 & 0.42 \\
Replay
& 43.62 & 3.88 & 33.44 & 1.22
& 60.95 & 1.14 & 53.94 & 0.46
& 61.96 & 1.96 & 57.38 & \textbf{0.39} \\
MoELoRA
& 43.92 & 4.86 & 32.62 & 1.55
& 60.89 & 1.43 & 54.27 & 0.69
& 61.78 & 1.38 & 57.50 & 0.43 \\
SMoLoRA
& 42.67 & 3.93 & 32.70 & 1.23
& 60.81 & 1.93 & 53.79 & 0.68
& 61.16 & 1.73 & 56.24 & 0.97 \\
Ours
& \textbf{44.38} & \textbf{3.05} & \textbf{34.15} & \textbf{0.47}
& \textbf{61.02} & 1.47 & \textbf{54.57} & \textbf{0.35}
& \textbf{61.99} & 2.02 & 57.61 & 0.51 \\
\bottomrule[0.7pt]
\end{tabular}

\end{table*}

\subsection{Additional Ablation Studies}
\label{app:hyperparameter_ablation}
We further analyze the sensitivity of StrLoRA to its main hyperparameters, including the number of LoRA experts $n$, the number of selected experts $k$, the expert rank $r$, and the regularization weight $\lambda$. As shown in Table~\ref{tab:additional_parameter_ablation}, the default setting $(n=4, k=2, r=16, \lambda=0.1)$ achieves the best overall balance, obtaining the highest MAP and the lowest MAF.

\begin{table*}[ht]
\centering
\caption{Additional ablation study of StrLoRA hyperparameters on InternVL3.5-8B.}
\label{tab:additional_parameter_ablation}
\setlength{\tabcolsep}{3.0pt}
\renewcommand{\arraystretch}{1.2}
\resizebox{\textwidth}{!}{%
\begin{tabular}{
c c c c
!{\vrule width 0.6pt}
*{8}{cc}
cc
!{\vrule width 0.6pt}
cc
}
\toprule
\multicolumn{4}{c}{\textbf{Hyperparameter}}
& \multicolumn{2}{c}{Places365}
& \multicolumn{2}{c}{RS}
& \multicolumn{2}{c}{AD}
& \multicolumn{2}{c}{VQAV2}
& \multicolumn{2}{c}{Fin}
& \multicolumn{2}{c}{Grounding}
& \multicolumn{2}{c}{OCRVQA}
& \multicolumn{2}{c}{TextCaps}
& \multicolumn{2}{c}{ImageNet}
& \multicolumn{2}{c}{Overall} \\
\cmidrule(r){1-4}
\cmidrule(r){5-6}
\cmidrule(lr){7-8}
\cmidrule(lr){9-10}
\cmidrule(lr){11-12}
\cmidrule(lr){13-14}
\cmidrule(lr){15-16}
\cmidrule(lr){17-18}
\cmidrule(lr){19-20}
\cmidrule(lr){21-22}
\cmidrule(l){23-24}
$n$ & $k$ & $r$ & $\lambda$
& AP$\uparrow$ & AF$\downarrow$
& AP$\uparrow$ & AF$\downarrow$
& AP$\uparrow$ & AF$\downarrow$
& AP$\uparrow$ & AF$\downarrow$
& AP$\uparrow$ & AF$\downarrow$
& AP$\uparrow$ & AF$\downarrow$
& AP$\uparrow$ & AF$\downarrow$
& AP$\uparrow$ & AF$\downarrow$
& AP$\uparrow$ & AF$\downarrow$
& MAP$\uparrow$ & MAF$\downarrow$ \\
\specialrule{0.08em}{0.12em}{0.08em}
4 & 2 & 16 & 0.1
& 33.37 & 2.99 & 69.65 & 0.75 & 36.27 & 3.92 & 70.11 & 1.13 & 75.95 & 0.38 & 57.36 & 2.53 & 53.00 & 3.29 & 84.48 & 0.74 & 76.49 & 1.46 & 61.85 & 1.91 \\
4 & 2 & 16 & 0.01
& 28.28 & 6.43 & 67.77 & 2.80 & 35.20 & 1.51 & 70.38 & 0.52 & 75.99 & 0.73 & 58.11 & 3.67 & 56.19 & 4.18 & 84.27 & 0.86 & 74.43 & 2.95 & 61.18 & 2.63 \\
4 & 2 & 8 & 0.1
& 26.28 & 11.01 & 67.17 & 2.55 & 31.47 & 2.51 & 70.14 & 0.87 & 72.42 & 0.53 & 55.90 & 4.05 & 57.38 & 4.78 & 83.44 & 1.02 & 66.28 & 4.81 & 58.94 & 3.57 \\
6 & 2 & 16 & 0.1
& 35.41 & 5.84 & 68.06 & 0.82 & 34.96 & 2.24 & 70.56 & 1.05 & 76.32 & 0.56 & 53.81 & 2.93 & 53.58 & 3.01 & 81.78 & 3.56 & 77.22 & 2.39 & 61.30 & 2.49 \\
6 & 3 & 16 & 0.1
& 33.11 & 6.21 & 67.79 & 0.82 & 34.96 & 2.24 & 70.36 & 1.05 & 75.66 & 0.57 & 55.97 & 2.83 & 57.12 & 2.83 & 82.37 & 3.54 & 74.82 & 2.52 & 61.35 & 2.51 \\
\bottomrule
\end{tabular}%
}

\end{table*}

\section{Case Study.}
\begin{figure*}[ht]
    \centering
    \includegraphics[width=0.95\textwidth]{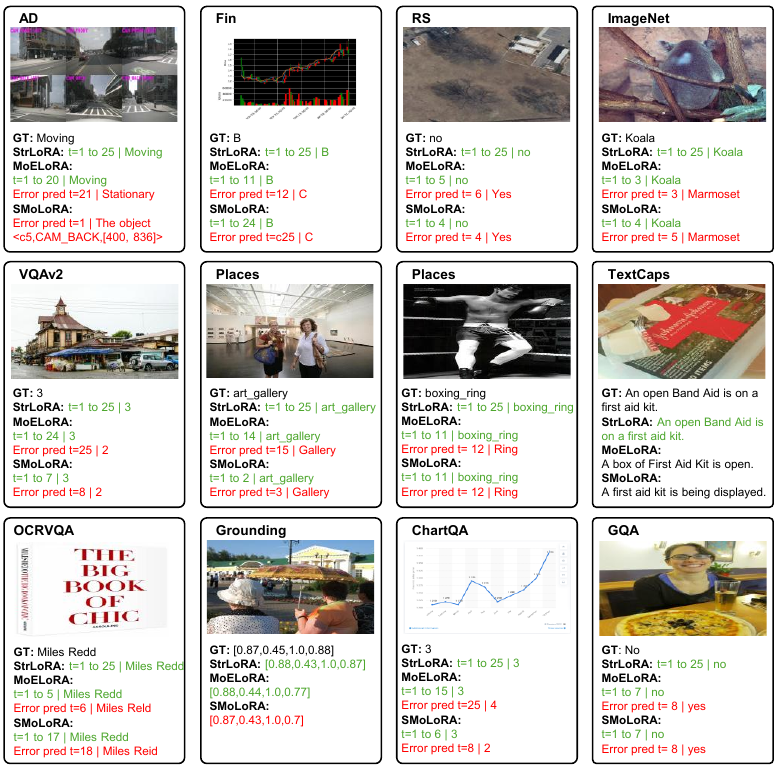}
    \caption{Qualitative temporal forgetting examples across the streaming process. Each panel tracks one fixed validation sample. Green text indicates correct predictions, and red text indicates the first observed erroneous prediction.
    }
    \label{fig:case_study}
\end{figure*}

We track the same validation samples across the streaming sequence. As shown in Fig.~\ref{fig:case_study}, StrLoRA keeps the correct prediction from $t=1$ to $t=25$ on diverse tasks, while MoELoRA and SMoLoRA often predict the sample correctly in early chunks but produce incorrect responses after later stream updates. These examples show that forgetting is not only reflected in aggregate accuracy drops, but also appears as concrete answer drift on individual samples.

\end{document}